\documentclass{article}

\usepackage{microtype}
\usepackage{graphicx}
\usepackage{booktabs} % for professional tables

\usepackage{amsmath}
\usepackage{amsfonts}
\usepackage{amssymb}
\usepackage{mathrsfs}
\usepackage{subcaption}
\usepackage{tabularx} % for aligned fixed width columns in table
\usepackage{placeins}

\allowdisplaybreaks

\usepackage{hyperref}

\newcommand{\erf}{\operatorname{erf}}
\newcommand{\argmin}{\arg\min}
\newcolumntype{L}[1]{>{\raggedright\arraybackslash}p{#1}}
\newcolumntype{C}[1]{>{\centering\arraybackslash}p{#1}}
\newcolumntype{R}[1]{>{\raggedleft\arraybackslash}p{#1}}

\usepackage[accepted]{icml2018}

\icmltitlerunning{Arbitrary Intra-Technique Transferability of Adversarial Examples for Logistic Regression}

\begin{document}

\twocolumn[
\icmltitle{Adversarial Perturbation Intensity Achieving Chosen Intra-Technique Transferability Level for Logistic Regression}



\icmlsetsymbol{equal}{*}

\begin{icmlauthorlist}
\icmlauthor{Martin Gubri}{ind}
\end{icmlauthorlist}

\icmlaffiliation{ind}{Independent Researcher, Bordeaux, France}

\icmlcorrespondingauthor{Martin Gubri}{martingubri@free.fr}

\icmlkeywords{Machine Learning, Adversarial Learning, Adversarial Examples, Logistic Regression, Asymptotic} %TODO : add ICML, for the submission

\vskip 0.3in
]



%TODO uncomment for ICML submission
\printAffiliationsAndNotice{}  % leave blank if no need to mention equal contribution

\begin{abstract}
Machine Learning models have been shown to be
vulnerable to adversarial examples, ie. the manipulation of data by a
attacker to defeat a defender's classifier at test time. We present a
novel probabilistic definition of adversarial examples in perfect or
limited knowledge setting using prior probability distributions on the
defender's classifier. Using the asymptotic properties of the logistic
regression, we derive a closed-form expression of the intensity of any
adversarial perturbation, in order to achieve a given expected
misclassification rate. This technique is relevant in a threat model of
known model specifications and unknown training data. To our knowledge,
this is the first method that allows an attacker to directly choose the
probability of attack success. We evaluate our approach on two real-world
datasets.
\end{abstract}

\section{Introduction}
\label{introduction}

Adversarial examples theory is the study of the strategies of a defender
and an attacker in the following threat model: an adversary has the
ability of modifying an input, noted \(x_0\) here, with the goal of
crafting a new input \(x_{adv}\) that will be misclassified by the
defender's classification model. The perturbation is noted
\(\delta_0\). Note that the attack happens at test time. The attacker
doesn't have the ability to alter the integrity of the model estimation.

\[ x_{adv} = x_0 + \delta_0 \]

With \(k\) known, untargeted adversarial examples crafting is usually
defined by the following optimization problem:

\[ \delta_0 = \argmin_{\delta}\lVert \delta \rVert _k \; \quad\text{s.t.}\quad f_d (x_0 + \delta ; \hat\theta_d) \neq y_0 \]

where $ f_d \colon \mathcal{X} \times \Theta \to \mathcal{Y}, \quad x \times \theta \mapsto y $ is the classifier~(or prediction function) of the defender's model, \(\hat\theta_d\) is the defender's estimate of the model parameters, \(\mathcal{X}\) the input space, \(\mathcal{Y}\) the output space, and \(\Theta\) the parameters space. Note that some authors prefer another definition for adversarial examples \cite{biggio_wild_2017}.

Some authors define the optimization problem by using
\(f_d(x_0 ; \hat\theta_d)\) instead of \(y_0\). We prefer to use
\(y_0\), because if the original input is already misclassified by the
model, ie. \(f_d(x_0 ; \hat\theta_d) \neq y_0\), then
\(\delta_0 = 0_p\), with \(p\) the number of features, if \(f_d\) and
\(\hat\theta_d\) are known. This point has consequences in limited
knowledge settings developed in part~4.

To this definition, the attacker may add application specific
constraints on \(\delta\). Various have been used in the literature:

\begin{itemize}
\item
  \(x_0 + \delta \in \mathcal{X}\), where \(\mathcal{X}\) may be
  \(\mathbb{N}\)
\item
  \(\lVert \delta \rVert _k < d_{\text{max}}\) \cite{biggio_evasion_2013}
\item
  \(\forall i, \delta_i \geq 0\) \cite{grosse_adversarial_2017}
\end{itemize}

Similar conditions can be derived only on a subset of features. For
simplicity reasons, we will not use any of these conditions in the
following. We assume that \(\mathcal{X} = \mathbb{R}^p\).

The attacker doesn't necessary have the same knowledge than the
defender. Knowing the defender's training data and model specifications,
the attacker can train an exact copy of \(f_d(\cdot, \hat\theta_d)\).
With partial knowledge of \(f_d(\cdot, \hat\theta_d)\), the attacker
can train a substitute model to craft adversarial examples from it \cite{biggio_evasion_2013,papernot_transferability_2016, papernot_practical_2017}. Papernot et~al. \yrcite{papernot_limitations_2016} build a typology of attacks depending on the knowledge that the attacker have on the defender's model and on the goal of the attacker. Biggio et~al. \yrcite{biggio_evasion_2013} detail the components of the attacker's knowledge:

\begin{itemize}
\item
  the defender's training data (completely or only a subset)
\item
  feature representation used by the defender
\item
  the type of learning algorithm and the decision function, that we
  called model specification
\item
  the defender's estimate of the model parameters.
\end{itemize}

The capability of the adversary may provide extra knowledge on the
defender's model. The typical example is the case of the attacker having
feedback from the defender's model \cite{papernot_practical_2017}.

The property of transferability of adversarial examples, defined by the
fact that some adversarial examples designed to fool a specific model
also fool other models, was observed by Goodfellow, Shlens, and Szegedy
\yrcite{goodfellow_explaining_2014}, Papernot et~al. \yrcite{papernot_transferability_2016}, Papernot et~al.  \yrcite{papernot_practical_2017}, among others. Papernot et~al. \yrcite{papernot_transferability_2016} identify two types of transferability: intra-technique transferability and cross-technique transferability.

The optimal L2-adversarial example for a logistic regression~(with
perfect knowledge) is the orthogonal projection of the example onto the
decision hyperplane \cite{moosavi-dezfooli_deepfool:_2016}. In
part~4, we use this technique to compute adversarial example, but our
method can be applied to any adversarial example crafting technique.

The intuition guiding our work our work is that an optimal adversarial example for the attacker surrogate model, given the limited knowledge of the attacker, may not achieve satisfactory intra-technique transferability. If the adversarial example is very close to the decision hyperplane, a very small difference between \(\hat\theta_d\) and \(\hat\theta_a\) can lead to failed attacks.

In part~2, we provide a probabilistic definition of an adversarial
example. In part~3, we recall the asymptotic properties of the logistic
regression. In part~4, we develop a closed-form approximate of an
adversarial example having a chosen expected successful attack rate, in
the threat model of known model specification but unknown training data
for binary classification by a logistic regression. In part~5, we apply
our method on 2 datasets.

\subsection*{Contributions}\label{contributions}

\begin{itemize}
\item
  We introduce a new probability-based definition of adversarial example
  having an arbitrary expected misclassification rate using prior
  distributions to formalize the attacker knowledge on the defender's
  classifier.
\item
  We make use of the asymptotic distribution of logistic regression
  parameters to derive a closed form method to craft adversarial
  examples having a chosen expected success attack rate, in a limited
  knowledge threat model. To our knowledge, this is the first method to
  allow an attacker to directly tune the probability of attack success.
\item
  We show that multiplying by the same scalar all adversarial
  perturbations of the test samples computed on the attacker's surrogate
  model may not be effective to improve intra-technique transferability.
\item
  We observe the importance of knowing the estimation method used by the
  defender, even for logistic regression.
\item
  We notice that in our setting the choice of the L2-regularization hyperparameter can be beneficial to the attacker by reducing the variance of parameters estimates.
\end{itemize}

\section{Probabilistic definition of adversarial example (\texorpdfstring{\boldmath$\alpha$} --adversarial example)}\label{probabilistic-definition} %TODO keep the parenthesis?

We define an \(\alpha\)-adversarial example as an adversarial example
with an expected rate of successful attacks of \(\alpha\) in an perfect
or imperfect knowledge setting:

\begin{multline}
\min_{\delta} \lVert \delta \rVert _k  \quad \text{s.t.} \\
\mathbb{P} \left( F_d ( x_0 + \delta ; \hat\theta_d ) \neq y_0 \Big\vert F_d \sim \mathscr{D}_{F_d} \cap \hat\theta_d \sim \mathscr{D}_{\hat\theta_d}(F_d) \right) \geq \alpha
\end{multline}

where \(\alpha \in [0,1]\) is chosen by the attacker.

\(F_d\) is a random function drawn from the sample space
\(\mathcal{F} \subset \mathcal{Y}^{\mathcal{X}}\) of the set of
prediction functions that the defender can use. \(\mathscr{D}_{F_d}\)
is the prior knowledge of the attacker on feature representation, model
type, its structural specifications (for example, the architecture of a
Neural Network), used by the defender. We consider \(\mathcal{X}\) as
the space of \emph{raw} data, and we include the data preprocessing step
into \(F_d\). \(\mathscr{D}_{\hat\theta_d}(F_d)\) captures the prior
knowledge on the defender's estimates of parameters, training data,
estimation methods, regularization, hyperparameters of the model and of
the feature representation. Then, \(\mathscr{D}_{\hat\theta_d}(F_d)\)
is the joined prior and hyperprior of the parameters of \(F_d\). The
Data Generating Process (DGP) \(\mathcal{P}_{\theta}(X, Y)\) is
parametrized by \(\theta\), the vector of true model parameters and
hyperparameters, and of feature representation hyperparameters. The
distribution of \(\hat\theta_d\) is conditioned by \(F_d\), because
the model (hyper)parameters may vary across model types.

We formalize the knowledge of the attacker using the joined probability
distribution of \(F_d\) and \(\hat\theta_d\). If the attacker knows
perfectly the true defender's decision function \(f_d\), then
\(\mathscr{D}_{F_d}\) is a deterministic distribution and
\(\forall f \in \mathcal{F}, \  \mathbb{P} \left( F_d = f \right) = 1 _{f_d}(f)\).
Instead the attacker may have only partial knowledge on the attacker
model. In practice, the attacker may know the state-of-the-art models or
the industry practices on a given task. Then, the adversary may be able
to draw a probability distribution on a set of models used by the
defender. The attacker may also draw probability distributions of
hyperparameters depending on the method used by the defender (random
search, grid search, etc.).

The attacker might want to estimate \(\mathscr{D}_{F_d}\) using
\(\hat f_a\) his/her estimate of \(f_d\) and estimate
\(\mathscr{D}_{\hat\theta_d}(F_d)\) using
\((\hat f_a, \hat\theta_a)\). This remark makes particularly sense if
the attacker has an oracle access to the defender model.

\section{Recalls of the asymptotic properties of the logistic regression} \label{recalls-asymptotic}

The logistic regression can be seen as a Generalized Linear Model (GLM)
with Binomial distribution and a logit link\footnote{The reader not familiar with the GLM theory can read McCullagh and Nelder \yrcite{mccullagh_generalized_1989}, which is the main book of reference on GLM but somewhat difficult, or Chapter~15 of Fox \yrcite{fox_applied_2016}  \href{https://www.sagepub.com/sites/default/files/upm-binaries/21121_Chapter_15.pdf}{available there}.}.

A GLM is defined by 3~components \citep[p.~379]{fox_applied_2016}:

\begin{enumerate}
\item
  A conditional distribution of the response variable \(Y_i\) given
  \(X_i\), member of the exponential family distribution. \(Y_i\) are
  independent.
\item
  A linear predictor,
  \(\eta_i = \beta_0 + \beta_1 X_{i,1} + \beta_2 X_{i,2} + \cdots + \beta_p X_{i,p}\).
\item
  A smooth and invertible link function \(g(\cdot)\),
  \(\eta_i = g(\mu_i)\).
\end{enumerate}

Note that the point~1. implies that a GLM is not only a transformation
of the classical linear model using a link function. GLM doesn't have
the hypothesis of normality of the residuals.

Then, the logistic regression is a special case of GLM with \(Y_i \sim \mathcal{B}(m_i, \pi_i)\) \footnote{The logistic regression can also be defined with \(Y_i\) following a Bernoulli distribution. Then,   \(\forall i \in [\![1;n]\!], m_i = 1\).} and the logit function as link (which is the canonical link of the Binomial distribution). Note that \(m_i\) is known, so it isn't a parameter of the model.

The Maximum Likelihood Estimator \(\hat\beta_{\text{MLE}}\) is
asymptotically normally distributed \citep[p.~121]{ferguson_course_1996}. It is
asymptotically unbiased with an asymptotic variance-covariance matrix
equals to the inverse of the Fisher information matrix

\[ 
	\left( X^{\intercal}WX \right)^{-1},
\]

with \(W\) a diagonal matrix of weights defined by \(W := \text{diag} \left\{ m_i \pi_i (1 - \pi_i) \right\}\)~\citep[p.~119]{mccullagh_generalized_1989}. \(W\) can be estimated by \(\widehat{W}=\text{diag} \left\{ m_i \hat\pi_i (1 - \hat\pi_i) \right\}\). Note that the same asymptotic property holds when \(n\) is fixed and \(m \to\infty\).

%\[ \sqrt{n} \left( \hat\beta_{\text{MLE}} - \beta \right) \xrightarrow[]{d} \mathcal{N}\left(0, \left( X^{\intercal}WX \right)^{-1} \right) \] % removed to save space

The ridge estimator in logistic regression, noted \(\hat\beta_{\text{L2}, \lambda_{\text{L2}}}\), is a maximum \emph{a posteriori} (MAP) estimator. Therefore it is asymptotically normal \citep[p.~140]{ferguson_course_1996}, asymptotically biased and the asymptotic variance-covariance matrix is given by \cite{le_cessie_ridge_1992}:

\[ \left( X^{\intercal}WX + 2 \lambda_{\text{L2}} I_p \right)^{-1} X^{\intercal}WX \left( X^{\intercal}WX + 2 \lambda_{\text{L2}} I_p \right)^{-1} \]

Then, if \(n\) is large, we can use the following approximations to
estimate the variance-covariance matrices:

\begin{equation*}
\begin{split}
    \widehat{\mathrm{Var}}\left( \hat\beta_{\text{MLE}} \right) =& \left( X^{\intercal}\widehat{W}X \right)^{-1} \\
    \widehat{\mathrm{Var}}\left( \hat\beta_{\text{L2}, \lambda_{\text{L2}}} \right) =& \left( X^{\intercal}\widehat{W}X + 2 \lambda_{\text{L2}} I_p \right)^{-1} X^{\intercal} \widehat{W}X \\
    &\left( X^{\intercal}\widehat{W}X + 2 \lambda_{\text{L2}} I_p \right)^{-1}
\end{split}
\end{equation*}

%\begin{eqnarray}
%    \widehat{\mathrm{Var}}\left( \hat\beta_{\text{MLE}} \right) &=& \left( X^{\intercal}\widehat{W}X \right)^{-1} \\
%    \widehat{\mathrm{Var}}\left( \hat\beta_{\text{L2}, \lambda_{\text{L2}}} \right) &=& \left( X^{\intercal}\widehat{W}X + 2 \lambda_{\text{L2}} I_p \right)^{-1} X^{\intercal} \widehat{W}X \left( X^{\intercal}\widehat{W}X + 2 \lambda_{\text{L2}} I_p \right)^{-1}
%\end{eqnarray}

\section{Approximation of \boldmath $\alpha$-adversarial examples for the logistic regression}\label{approximation-alpha-adversarial-examples}

For convenience, in this section we define \(\tilde x_0 :=~\begin{pmatrix}1\\x_0\end{pmatrix}~
 \in~\mathbb{R} ^{p+1}\) and \(\tilde \delta_0 := \begin{pmatrix}0\\ \delta_0\end{pmatrix}
 \in \mathbb{R} ^{p+1}\).

We will consider the following \textit{threat model}:

\textbf{Perfect knowledge of \boldmath \(f_d\)}
\begin{itemize}
  \item
    The defender is using a logistic regression to perform a binary
    classification task,
%    \begin{multline*}
%    \forall (x_0, \beta) \in \mathbb{R} ^p \times \mathbb{R} ^{p+1}, \\
%    f_d(x_0, \beta) =
%    \begin{cases} 1 & \text{if } \tilde x_0 \beta > 0, \\
%    0 & \text{otherwise}.
%    \end{cases}
%    \end{multline*}
	$$
    \forall (x_0, \beta) \in \mathbb{R} ^p \times \mathbb{R} ^{p+1}, \ 
    f_d(x_0, \beta) =
    \begin{cases} 1 & \text{if } \tilde x_0 ^\intercal \beta > 0, \\
    0 & \text{otherwise}.
    \end{cases}
    $$
    
    \item
    The attacker knows perfectly the defender's feature representation.
\end{itemize}

\textbf{Limited knowledge of \boldmath \(\hat\theta_d\)}
\begin{itemize}
  \item
    The attacker doesn't have access to the defender's training data.
  \item
    The attacker has access to some surrogate training data generated by
    the same Data Generating Process (DGP) parametrized by
    \(\beta \in \mathbb{R}^{p+1}\).
  \item
    The attacker knows the specifications of the logistic regression
    (regularization method and hyperparameters, estimation method).
\end{itemize}

\medskip

Therefore the defender's parameters estimation \(\hat\beta_d\) is fixed but unknown by the attacker. The attacker can compute \(\hat\beta_a\) using his/her own data.

The goal of the attacker is to find \(\delta^*\) solving problem \ref{eq:min_delta_star}.

\begin{equation}\label{eq:min_delta_star}
\delta^* = \argmin_{\delta \in \mathbb{R}^p} \lVert \delta \rVert _k \; \quad\text{s.t.}\quad \mathbb{P} \left[ f_d(x_0 + \delta ; \hat\beta_d) \neq y_0 \right] \geq \alpha 
\end{equation}

where \(k \in \{1,2\}\) and \(\alpha \in [0,1]\) are chosen by the
attacker.

For simplicity purposes, we consider the following suboptimal problem.

\textbf{First step:} The attacker compute an adversarial example for \emph{his/her own model}. Any adversarial example crafting technique can be
used.

\[\delta_0 = \argmin_{\delta \in \mathbb{R}^p} \lVert \delta \rVert _k \; \quad\text{s.t.}\quad f_d(x_0 + \delta ; \hat\beta_a) \neq y_0\]

In the following, we consider the L2-optimal adversarial example ($k=2$) which is the orthogonal projection of \(x_0\) on the decision hyperplane \(\mathcal{H}_a = \{ x \in \mathbb{R}^p \; | \; f_d(x ; \hat\beta_a) = 0 \} \) \cite{moosavi-dezfooli_deepfool:_2016}. The associated perturbation can be computed by

\[ \delta_0 = - \frac{ \tilde x_0^\intercal \hat\beta _a}{ \lVert \hat\beta _{\text{a;-0}} \rVert^2_2 }\hat\beta _{\text{a;-0}}
\]

where \( \hat\beta _{\text{a;-0}} = \left(\hat\beta_{a;1}, \hat\beta_{a;2}, \cdots , \hat\beta_{a;p} \right)^\intercal\).

\textbf{Second step:} The attacker searches an optimal scalar
\(\lambda^*\), the intensity of the adversarial perturbation
\(\delta_0\), needed to achieve an expected misclassification rate on
the defender's model of at least \(\alpha\):

\[ \lambda^* = \argmin_{\lambda \in \mathbb{R}} \lVert \lambda \delta_0 \rVert _{k'} \; \  \text{s.t.} \  \mathbb{P} \left[ f_d(x_0 + \lambda \delta_0 ; \hat\beta_d) \neq y_0 \right] \geq \alpha \]

with \(k' \in \{1,2\}\). It can be simplified as

\begin{equation}\label{eq:suboptimal_problem_definition}
\lambda^* = \argmin_{\lambda \in \mathbb{R}} \lambda^2 \; \quad\text{s.t.}\quad \mathbb{P} \left[ f_d(x_0 + \lambda \delta_0 ; \hat\beta_d) \neq y_0 \right] \geq \alpha
\end{equation}

We denote the \(\alpha\)-adversarial example: \(x_{adv}^* := x_0 + \lambda^* \delta_0\).

Problem \ref{eq:suboptimal_problem_definition} can be rewritten as

\[ \min_{\lambda \in \mathbb{R}} g(\lambda) \enspace\text{s.t.}\enspace h(\lambda) \geq 0 \]
with $ g \colon \mathbb{R} \to \mathbb{R}^+, \quad \lambda \mapsto \lambda^2 $ and  $ h \colon \mathbb{R} \to [-1,1], \quad$ \allowbreak $\lambda \mapsto \mathbb{P} \left[ f_d(x_0 + \lambda \delta_0 ; \hat\beta_d) \neq y_0 \right] - \alpha $.

\(g\) and \(h\) are of class \(C^1\). Then using the complementary
slackness of the Karush--Kuhn--Tucker conditions, if \(\lambda^*\) is a
local optimum, \(h(\lambda^*) g'(\lambda^*)=0\) . If \(h(\lambda^*) = 0\) the constraint is said to be saturated, and if \(g'(\lambda^*) = 0\) it is not saturated.

\subsection{Case 1: Constraint saturated, ie. \boldmath $\mathbb{P} \left[ f_d(x_0 + \lambda \delta_0 ; \hat\beta_d) \neq y_0 \right] = \alpha$}\label{case-1-constraint-saturated}

For convenience, we define the random variable \(Z\) as follow:

\[ \forall \lambda \in \mathbb{R}, \quad Z := (\tilde x_0 + \lambda \tilde \delta_0)^\intercal \hat\beta_d \]

For large samples,
\(\hat\beta_d \rightsquigarrow \mathcal{N}\left( \mathbb{E}(\hat\beta_d), \mathrm{Var}(\hat\beta_d) \right)\).

Then, \(Z \rightsquigarrow \mathcal{N}\left( \mathbb{E}(Z), \mathrm{Var}(Z) \right) \) with

\begin{align*} 
\mathbb{E}(Z) &= (\tilde x_0 + \lambda \tilde \delta_0)^\intercal \mathbb{E}(\hat\beta_d) ,\\ 
\mathrm{Var}(Z) &= (\tilde x_0 + \lambda \tilde \delta_0)^\intercal \mathrm{Var}(\hat\beta_d)(\tilde x_0 + \lambda \tilde \delta_0).
\end{align*}

The attacker estimates \(\mathbb{E}(\hat\beta_d)\) by \(\hat\beta_a\), and   \(\mathrm{Var}(\hat\beta_d)\) by \(\widehat{\mathrm{Var}}(\hat\beta_a)\) which is computed as explained in part~\ref{recalls-asymptotic}.

\subsubsection{Subcase a: \(y_0 = 1\)} \label{subcase-a}

Using the quantile function of the normal distribution, if \(y_0 = 1\):

\begin{align}
  \nonumber
  &\mathbb{P} \left[ f_d(x_0 + \lambda \delta_0 ; \hat\beta_d) \neq y_0 \right] = \alpha \\
  \nonumber
  & \Leftrightarrow \mathbb{P} \left( Z \leq 0 \right) = \alpha \\
  \nonumber
  & \Leftrightarrow (\tilde x_0 + \lambda \tilde \delta_0)^\intercal \hat\beta_d \\
  \nonumber
  & \quad + \sqrt{(\tilde x_0 + \lambda \tilde \delta_0)^\intercal \mathrm{Var}(\hat\beta_d) (\tilde x_0 + \lambda \tilde \delta_0) } \sqrt{2} \erf^{-1}(2 \alpha -1) \\
  & \quad = 0 \label{eq:condition_lambda_star}
\end{align}

Equation \ref{eq:condition_lambda_star} is estimated by the attacker by~:

\begin{align*}
	&(\tilde x_0 + \lambda \tilde \delta_0)^\intercal \hat\beta_a \\
	& \quad + \sqrt{(\tilde x_0 + \lambda \tilde \delta_0)^\intercal \widehat{\mathrm{Var}}(\hat\beta_a) (\tilde x_0 + \lambda \tilde \delta_0) } \sqrt{2} \erf^{-1}(2 \alpha -1) \\
	& \quad = 0 \\
	& \Rightarrow \left[ (\tilde x_0 + \lambda \tilde \delta_0)^\intercal \hat\beta_a \right]^2 \\
	& \quad - (\tilde x_0 + \lambda \tilde \delta_0)^\intercal \widehat{\mathrm{Var}}(\hat\beta_a) (\tilde x_0 + \lambda \tilde \delta_0) 2 \left[ \erf^{-1}(2 \alpha -1) \right]^2 \\
	& \quad = 0 \\
	& \Rightarrow (\tilde x_0 + \lambda \tilde \delta_0)^\intercal \hat\beta_a \hat\beta_a^\intercal (\tilde x_0 + \lambda \tilde \delta_0) \\
	& \quad - (\tilde x_0 + \lambda \tilde \delta_0)^\intercal \widehat{\mathrm{Var}}(\hat\beta_a) (\tilde x_0 + \lambda \tilde \delta_0) 2 \left[ \erf^{-1}(2 \alpha -1) \right]^2 \\
	& \quad = 0 \\
	& \Rightarrow (\tilde x_0 + \lambda \tilde \delta_0)^\intercal \left[ \hat\beta_a \hat\beta_a^\intercal - 2 \left( \erf^{-1}(2 \alpha -1) \right)^2 \widehat{\mathrm{Var}}(\hat\beta_a) \right] \\
	& \quad (\tilde x_0 + \lambda \tilde \delta_0) = 0 \\
	& \Rightarrow \tilde x_0^\intercal A \tilde x_0 + \lambda( \tilde x_0^\intercal A \tilde \delta_0 + \tilde \delta_0^\intercal A \tilde x_0 ) + \lambda^2 \tilde \delta_0^\intercal A \tilde \delta_0 = 0 
\end{align*}

with \(A := \hat\beta_a \hat\beta_a^\intercal - 2 \left( \erf^{-1}(2 \alpha -1) \right)^2 \widehat{\mathrm{Var}}(\hat\beta_a)\).

Then, \(\lambda \in \mathbb{R}\) can be computed by solving a second degree equation. If there are two solutions, we choose the one that satisfy Equation \ref{eq:condition_lambda_star}. We denote the solution of the second degree equation \(\lambda_\alpha\).

\subsubsection{Subcase b: \(y_0 = 0\)} \label{subcase-b}

\begin{align*} 
	&\mathbb{P} \left[ f_d(x_0 + \lambda \delta_0 ; \hat\beta_d) \neq y_0 \right] = \alpha \\
	& \Leftrightarrow \mathbb{P} \left( Z \leq 0 \right) = 1-\alpha \\
\end{align*}

\(\lambda_\alpha\) can be derived similarly to subcase a, replacing
\(\alpha\) by \(1-\alpha\).

\subsection{Case 2: Constraint not saturated, ie. \boldmath $\mathbb{P} \left[ f_d(x_0 + \lambda \delta_0 ; \hat\beta_d) \neq y_0 \right] > \alpha $} \label{case-2-constraint-not-saturated}

In this case, \(g'(\lambda^*) = 0\). It immediately follows that
\(\lambda^* = 0\).

\bigskip

If \(\mathbb{P} \left[ f_d(x_0 ; \hat\beta_d) \neq y_0 \right] > \alpha\), then 0 is the global minimum. Otherwise and if \(\lambda_\alpha\) exists, then \(\lambda_\alpha\) is the global minimum, because it is the unique point that saturates the constraint.

To sum up, the problem \ref{eq:suboptimal_problem_definition} can be solved by:

\[
\lambda^* = \begin{cases} 0 & \text{if } \mathbb{P} \left[ f_d(x_0 ; \hat\beta_d) \neq y_0 \right] > \alpha, \\
\lambda_\alpha & \text{otherwise}.\end{cases}
\]

\section{Applications}\label{applications}

We applied our analysis on 2 datasets: the UCI spambase set and the dogs vs cats image set. These two datasets are binary classification problems. The code is available for reproducibility purpose on \href{https://github.com/Framartin/adversarial-logistic}{GitHub} and \href{https://framagit.org/martin.gubri/ext_adv_ex}{Framagit}. %TODO: anonimized

\begin{table}[tb]
\caption{Classification accuracies for the 3 estimation methods studied on the spam dataset.}
\label{table:accuracy_spam}
\vskip 0.15in
\begin{center}
\begin{small}
\begin{sc}
\begin{tabular}{L{0.31\columnwidth}|C{0.25\columnwidth}C{0.28\columnwidth}}
\toprule
Estimation method   & Accuracy \newline in-sample  & Accuracy \newline out-of-sample \\
\midrule
IRLS                    & 93.11\%  & 92.61\%  \\
Unregularized liblinear   & 93.11\%  & 92.75\%  \\
L2-regularized liblinear  & 93.14\%  & 92.75\%  \\
\bottomrule
\end{tabular}
\end{sc}
\end{small}
\end{center}
\vskip -0.1in
\end{table}

\subsection{Spambase Data Set}\label{spambase-data-set}

The \href{https://archive.ics.uci.edu/ml/datasets/spambase}{UCI spam dataset} is small enough to estimate the logistic regression using the Iteratively Reweighted Least Squares (IRLS) estimation method, generally used for GLM, provided by the statsmodels Python module. We also trained an unregularized and a L2-regularized logistic regression using the Scikit-learn implementation and the liblinear solver.

The accuracies are pretty similar between estimation methods (Table \ref{table:accuracy_spam}). But the estimated variance-covariance matrices are very different from the IRLS estimation and the liblinear ones. It leads to very different intensities to achieve the same misclassification level for some examples. In Table \ref{table:spam_example_0}, we can observe very different values of $\lambda^*$ across the 3 estimation methods studied for a arbitrary example $x_0$ in the test set. This is mainly due to the high difference in the estimations of \(\beta_{41}\), because \(\hat\beta_{41; \text{IRLS}}=-48.08\), \(\hat\beta_{41; \text{URLB}} = -3.07\) and \(\widehat{\mathrm{Var}}\left( \hat\beta_{41; \text{IRLS}} \right) - \widehat{\mathrm{Var}}\left( \hat\beta_{41; \text{URLB}} \right) = 1332.74\) whereas the second biggest element-wise difference between the two covariance matrices in absolute value is \(12.56\). It emphasis the importance of knowing the estimation method used by the defender.

\begin{table*}[tb]
\caption{Predicted probabilities of being a spam by the attacker model of an arbitrary test examples \(x_0\), its original perturbation \(x_0 + \delta_0\), and its intensified perturbation \(x_0 + \lambda^* \delta_0\), and values of the intensities \(\lambda^*\), for the 3 estimation methods studied on the spam dataset. Note that $y_0 = 1$ and $\alpha=0.95$.}
\label{table:spam_example_0}
\vskip 0.15in
\begin{center}
\begin{small}
\begin{sc}
\begin{tabular}{L{0.5\columnwidth}|C{0.3\columnwidth}C{0.3\columnwidth}R{0.3\columnwidth}}
\toprule
 Predicted probabilities and Intensities & IRLS  & Unregularized liblinear &  L2-regularized liblinear \\
\midrule
$\mathbb{P}(Y=1 | X=x_0, \hat\beta_a)$                         & 99.999986\%         & 99.999892\% & 99.999919\% \\
$\mathbb{P}(Y=1 | X=x_0 + \delta_0, \hat\beta_a)$              & 49.999606\%         & 49.999656\% & 49.999649\% \\
$\mathbb{P}(Y=1 | X=x_0 + \lambda^* \delta_0, \hat\beta_a)$ & $1.687831e^{-74}\%$ & 1.004755\%  & 2.040357\%  \\
$\lambda^*$                                                    & 12.059536           & 1.333934    & 1.275921    \\
\bottomrule
\end{tabular}
\end{sc}
\end{small}
\end{center}
\vskip -0.1in
\end{table*}

%\begin{figure*}[ht]
%\vskip 0.2in
%\begin{center}
%\begin{subfigure}[t]{\columnwidth} %{0.45\textwidth}
%	\centerline{\includegraphics[width=\columnwidth]{images/spam_intensities_alphas.png}}
%	\caption{\label{subfig:spam_intensity_vs_alpha_a}}
%\end{subfigure}
%\begin{subfigure}[t]{\columnwidth} %{0.45\textwidth}
%	\centerline{\includegraphics[width=\columnwidth]{images/spam_intensities_alphas_zoom.png}}
%	\caption{\label{subfig:spam_intensity_vs_alpha_b}}
%\end{subfigure}
%\caption{Intensities of perturbations versus misclassification levels for the same example
%\(x_0\) than table \ref{table:spam_example_0}. Figure~\ref{subfig:spam_intensity_vs_alpha_b} is zoomed in for better visualization.}
%\label{fig:spam_intensity_vs_alpha}
%\end{center}
%\vskip -0.2in
%\end{figure*}

%\begin{figure*}[ht]
%\vskip 0.2in
%\begin{center}
%\begin{subfigure}[t]{\columnwidth} %{0.45\textwidth}
%	\centerline{\includegraphics[width=\columnwidth]{images/spam_violinplot.pdf}}
%	\caption{\label{subfig:spam_violinplot_a}}
%\end{subfigure}
%\begin{subfigure}[t]{\columnwidth} %{0.45\textwidth}
%	\centerline{\includegraphics[width=\columnwidth]{images/spam_violinplot_zoom.pdf}}
%	\caption{\label{subfig:spam_violinplot_b}}
%\end{subfigure}
%\caption{Violin plot of the intensities of perturbations in the test set for \(\alpha=0.9\). Figure~\ref{subfig:spam_violinplot_b} is zoomed in for better visualization.}
%\label{fig:spam_violinplot}
%\end{center}
%\vskip -0.2in
%\end{figure*}

\begin{figure}[tb]
\vskip 0.2in
\begin{center}
\centerline{\includegraphics[width=\columnwidth]{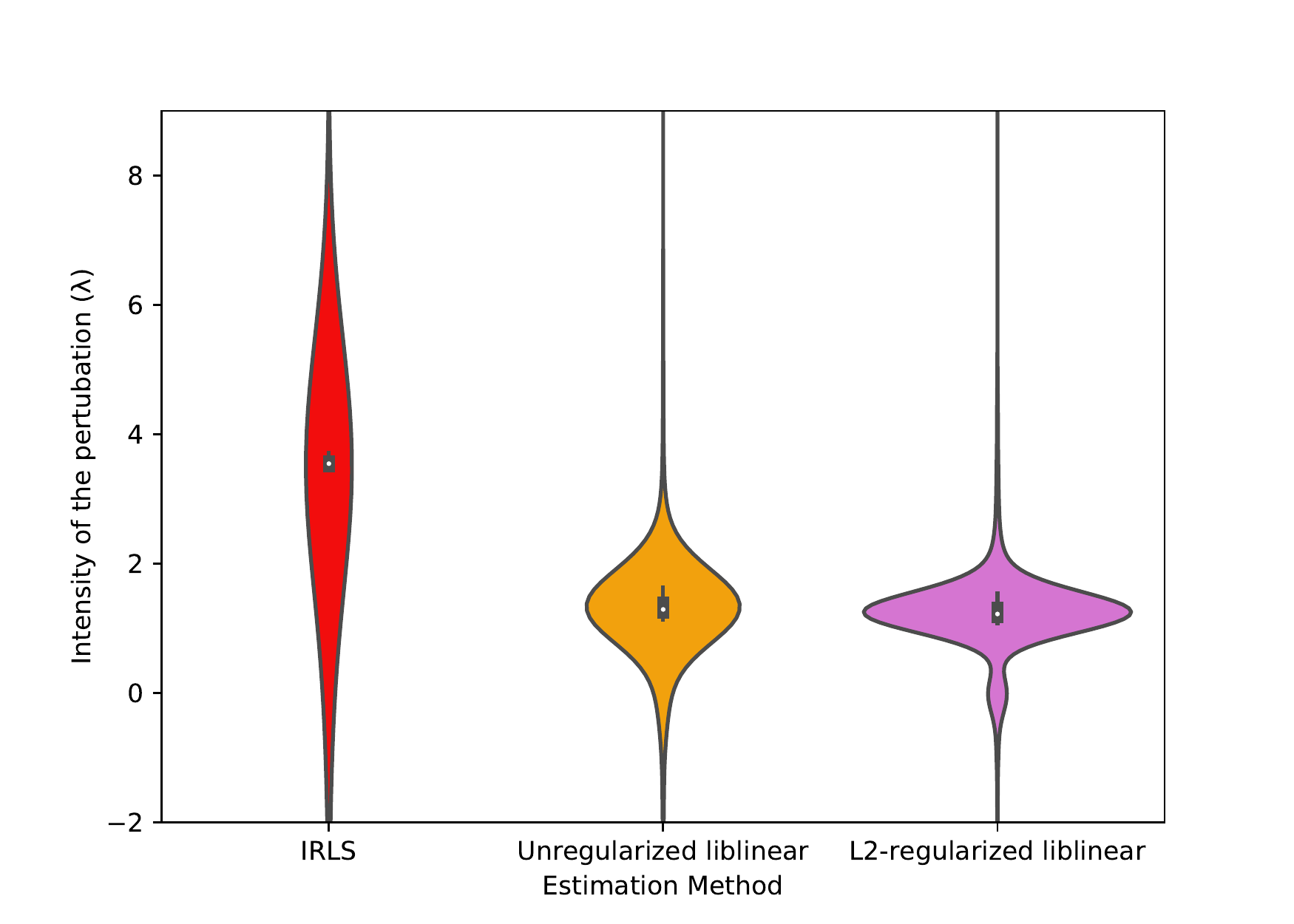}}
\caption{Violin plot of the intensities of perturbations in the test set for \(\alpha=0.9\). The figure is zoomed in for better visualization.}
\label{fig:spam_violinplot}
\end{center}
\vskip -0.2in
\end{figure}

\begin{figure}[tb]
\vskip 0.2in
\begin{center}
	\centerline{\includegraphics[width=\columnwidth]{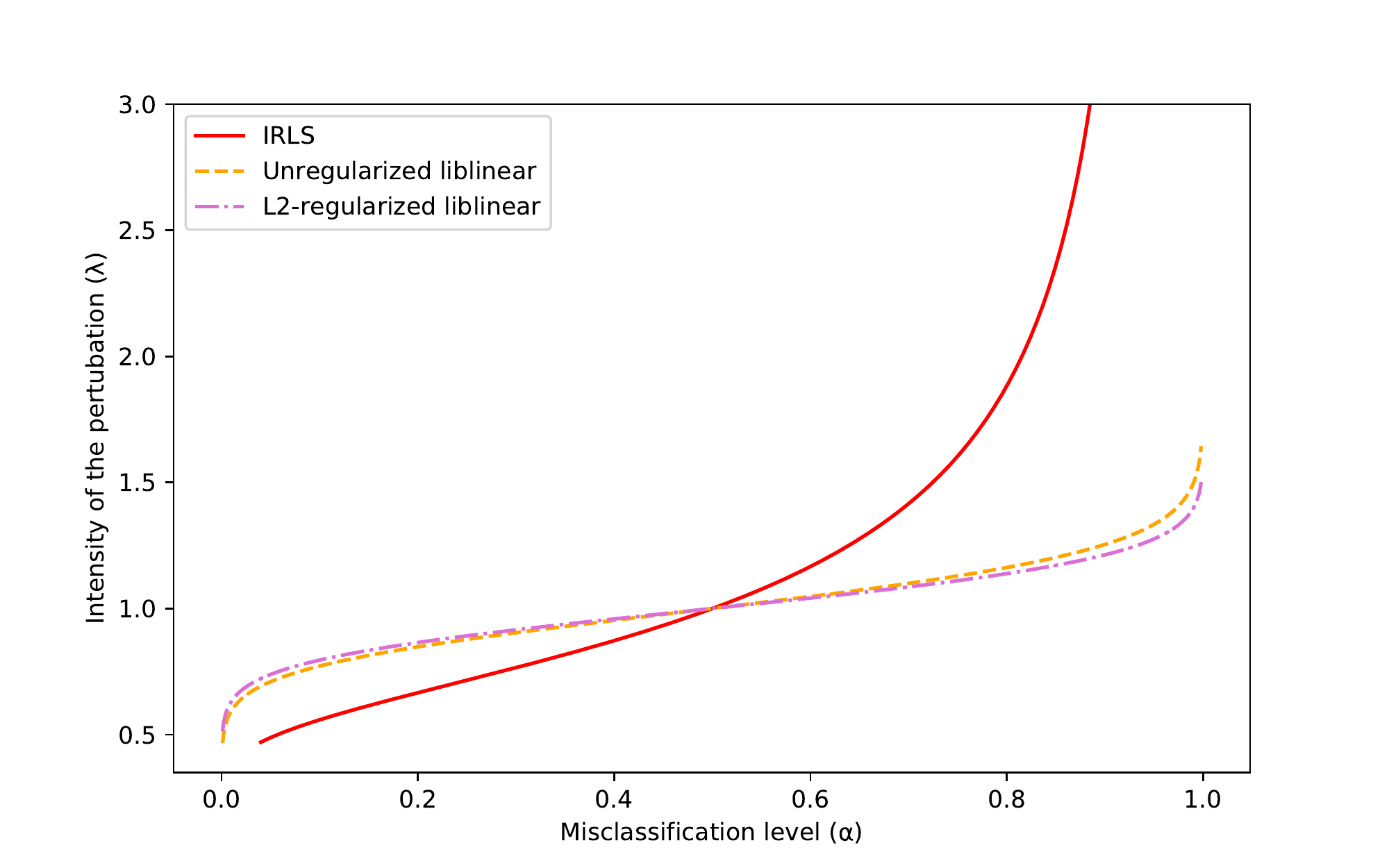}}
	\caption{Intensities of perturbations versus misclassification levels for the same example
\(x_0\) than Table \ref{table:spam_example_0}.}
\label{fig:spam_intensity_vs_alpha}
\end{center}
\vskip -0.2in
\end{figure}

Table \ref{table:spam_example_0} also reports the estimated probabilities of being a spam by the attacker model for the intensified perturbations of $x_0$ across estimation methods. It insists on the fact that we do not have to confound the estimated probability of an adversarial example to be in a specific class by the attacker's surrogate model, and the probability $\alpha$ of being classified in a specific class by the defender's model.

Figure \ref{fig:spam_intensity_vs_alpha} represents the intensity of the adversarial orthogonal perturbation against misclassification levels from 0 to 1 for the same arbitrary example $x_0$. The intensity associated to the IRLS estimation is higher than the other two for all values of $\alpha$ higher than 0.5, and it explodes sooner when $\alpha$ tends towards 1. Notice that the the intensities have different scale across examples. It confirm our intuition that multiplying all examples by the same scalar in not the best way to improve intra-technique transferability.

Figure \ref{fig:spam_violinplot} shows the box plots and the kernel density estimations of the perturbation intensity in the test set for a fixed misclassification level of \(0.90\) across the estimation methods. The median of intensities computed on the unregularized model is sightly greater than the one on the L2-regularized model.

\begin{figure}[tb]
\vskip 0.2in
\begin{center}
\centerline{\includegraphics[width=\columnwidth]{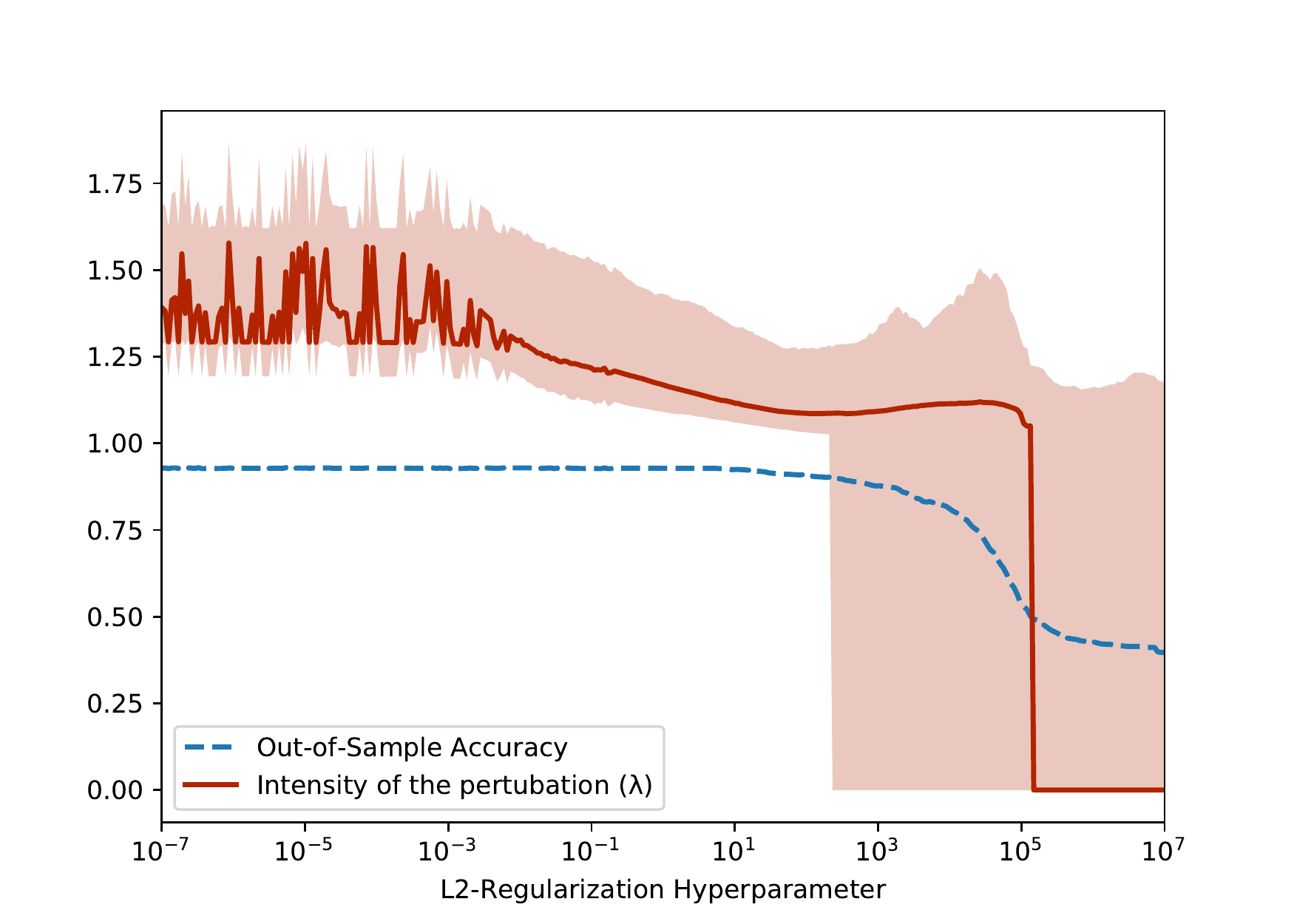}}
\caption{Quantiles of the empirical distributions of the intensities $\lambda^*$ in the test set and out-of-sample accuracy, versus values of the L2-regularization hyperparameter $\lambda_{\text{L2}}$ (in log-scale). The red lower bound represents the first decile, the red upper one the last decile, and the red line the median.}
\label{fig:spam_regularization}
\end{center}
\vskip -0.2in
\end{figure}

Figure \ref{fig:spam_regularization} represents the evolution of the intensities of the adversarial perturbations with respect to the L2-regularization hyperparameter. The relation is not straightforward. A very strong regularization is beneficial to the attacker, because it leads to very small parameter variance, which at the end implies smaller perturbations to achieve the same misclassification level. A very small values of regularization leads to instability in the intensities, which is good for the defender. Interestingly, when the regularization is strong enough to lower the accuracy, it increases the intensity until saturation of the constraint ($\lambda^* = 0$). With this exception in mind, we can globally said that if regularization leads to better estimates in terms of MSE \cite{mansson_ridge_2011}, it is also beneficial to the attacker. Then, we make the hypothesis that there is a trade-off in the defender's choice of L2-regularization hyperparameter between performance and security.

\subsection{Dogs vs Cats Images} \label{dogs-vs-cats-images}

We also applied our results to the Dogs versus Cats images dataset, available on \href{https://www.kaggle.com/c/dogs-vs-cats/data}{Kaggle}, which is composed of 25000 labeled images of cats and dogs. 

We preprocess the images by normalizing the luminance and resizing them to a squared shape of 64 by 64 pixels. The low resolution is necessary for us, because the computation of the variance-covariance matrix needs the inversion of a \(p \times p\) matrix. We preserve the aspect ratio by adding gray bars as necessary to make them square. Even using 64 by 64 pixels images, the resulting 12288 features are too large for the GLM estimation using IRLS. We only used the Scikit-learn implementation of logistic regression. We trained a L2-regularized logistic regression, using the SAG solver, where the regularization hyperparameter is chosen by grid search of 100 values on 3-Fold Cross Validation.

\begin{table}[tb]
\caption{Classification accuracies the logistic regression trained on the dogs vs cats images.}
\label{table:accuracy_cats}
\vskip 0.15in
\begin{center}
\begin{small}
\begin{sc}
\begin{tabular}{lcccr}
\toprule
   & Accuracy \\
\midrule
In-sample     & 73.37\% \\
Out-of-sample & 58.07\% \\
\bottomrule
\end{tabular}
\end{sc}
\end{small}
\end{center}
\vskip -0.1in
\end{table}

The accuracy of our model is poor (Table \ref{table:accuracy_cats}), because the data are not linearly separable. It clearly overfits our training data.

\begin{figure*}[p]
\vskip 0.2in
\begin{center}
\begin{subfigure}[t]{0.90\columnwidth}
	\centerline{\includegraphics[width=\columnwidth]{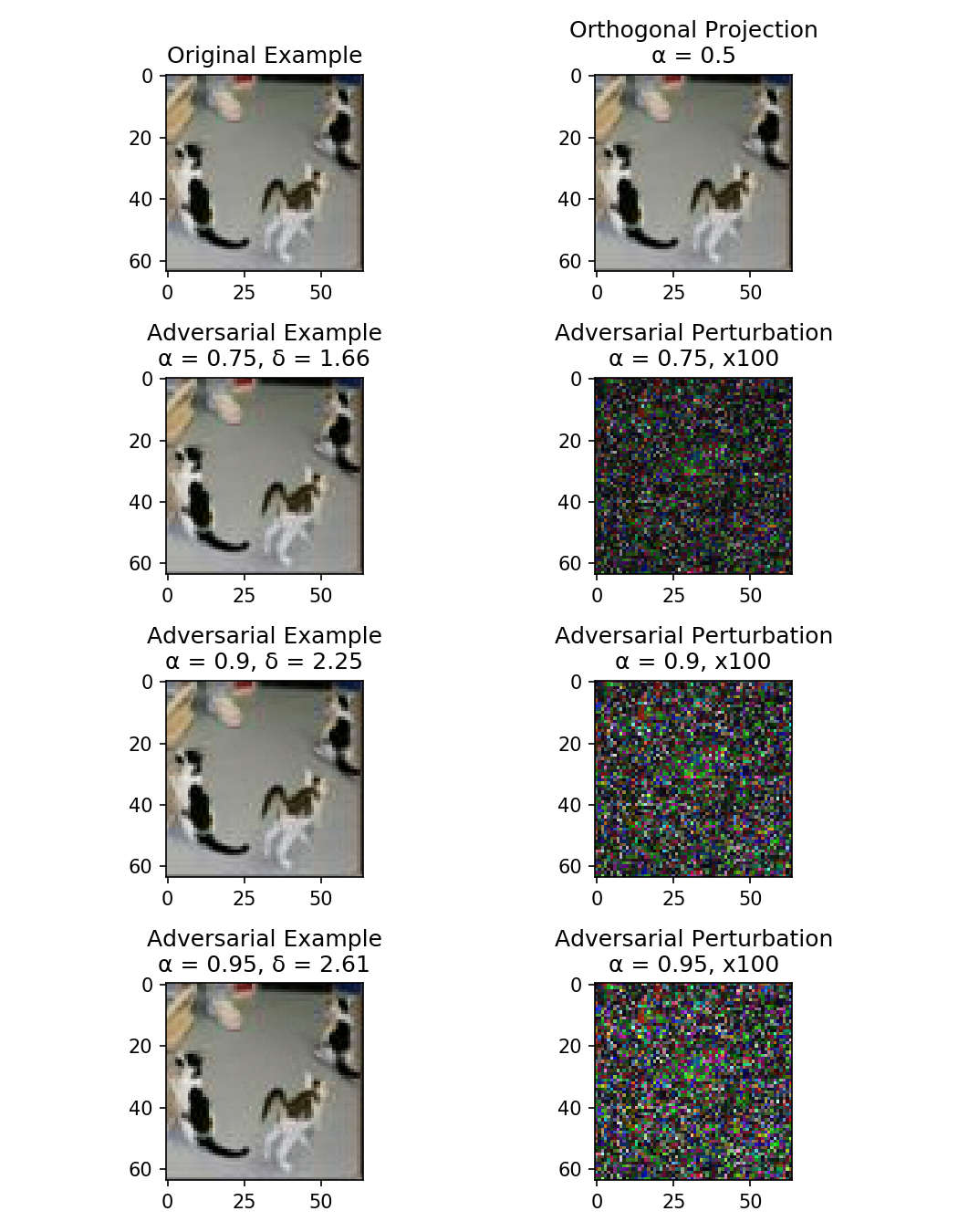}}
	\caption{Image 1}\label{subfig:cats_adv_test_a}
\end{subfigure}
\begin{subfigure}[t]{0.90\columnwidth}
	\centerline{\includegraphics[width=\columnwidth]{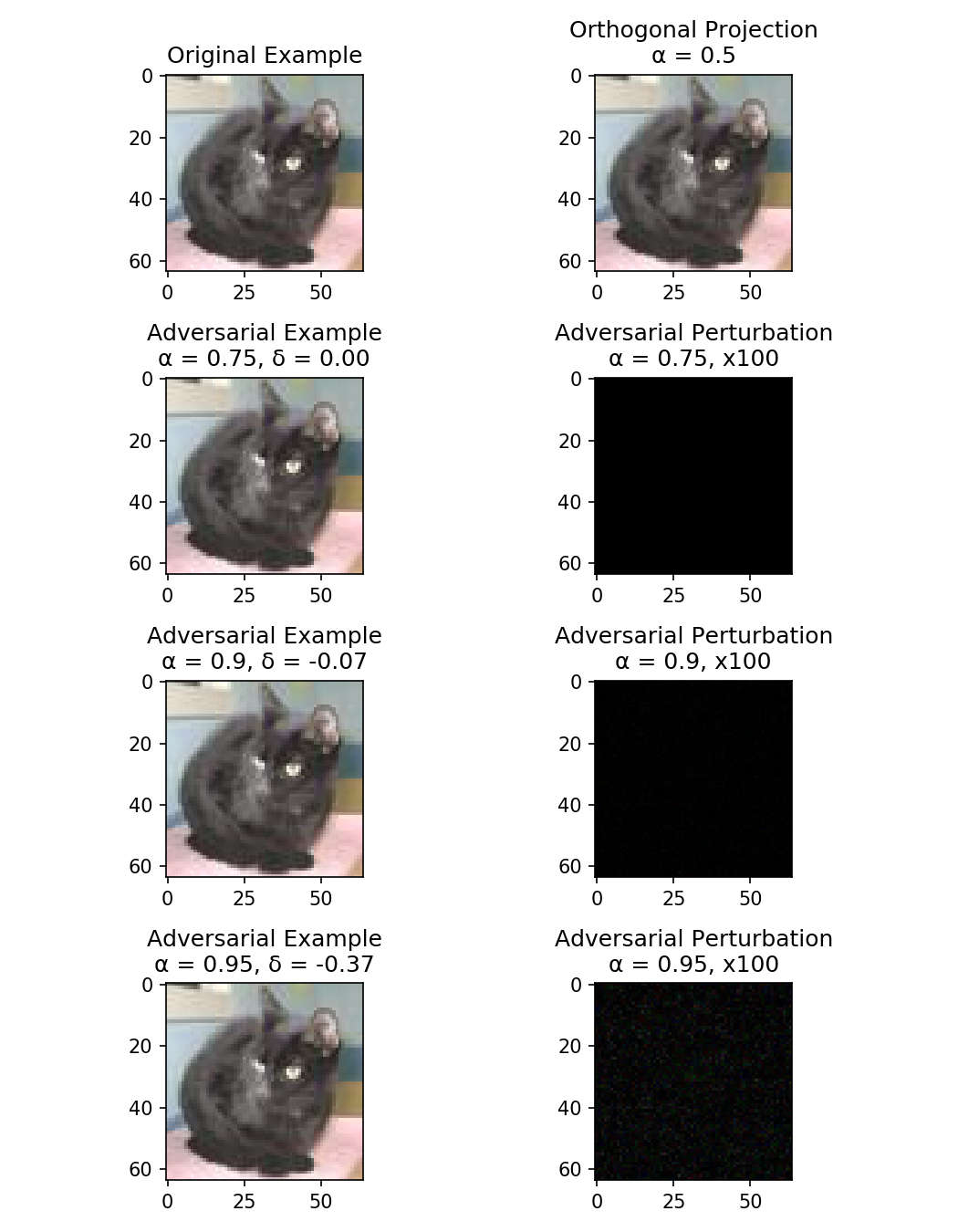}}
	\caption{Image 2}\label{subfig:cats_adv_test_b}
\end{subfigure}
\caption{Original image, attacker's L2-optimal adversarial image, adversarial images achieving misclassification levels of 0.75, 0.90 and 0.95, and their associated perturbations, for 2 images from the test set. Image 1 is correctly classified by our logistic regression and Image 2 is not. Perturbations are represented in absolute values and multiplied by 100.}
\label{fig:cats_adv_test}
\end{center}
\vskip -0.2in
\end{figure*}

\begin{figure*}[p]
\vskip 0.2in
\begin{center}
\begin{subfigure}[t]{\columnwidth} %{0.45\textwidth}
	\centerline{\includegraphics[width=\columnwidth]{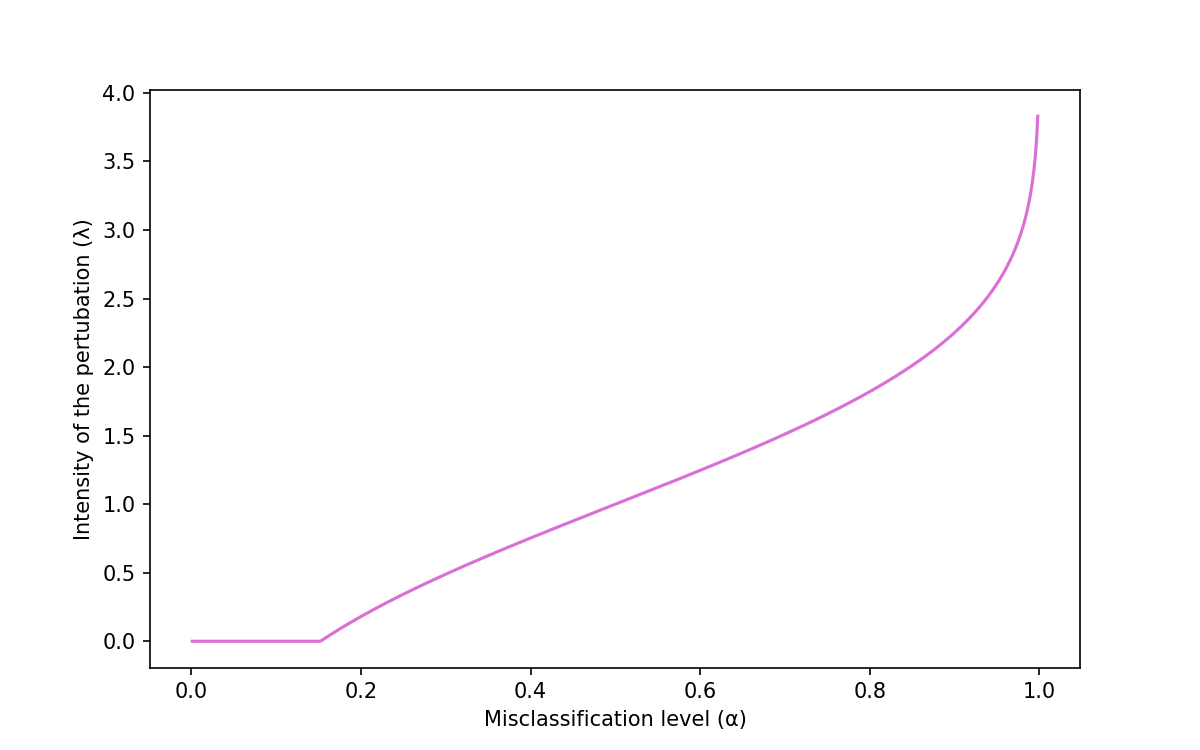}}
	\caption{Test Image 1}\label{subfig:cats_intensity_vs_alpha_test_a}
\end{subfigure}
\begin{subfigure}[t]{\columnwidth} %{0.45\textwidth}
	\centerline{\includegraphics[width=\columnwidth]{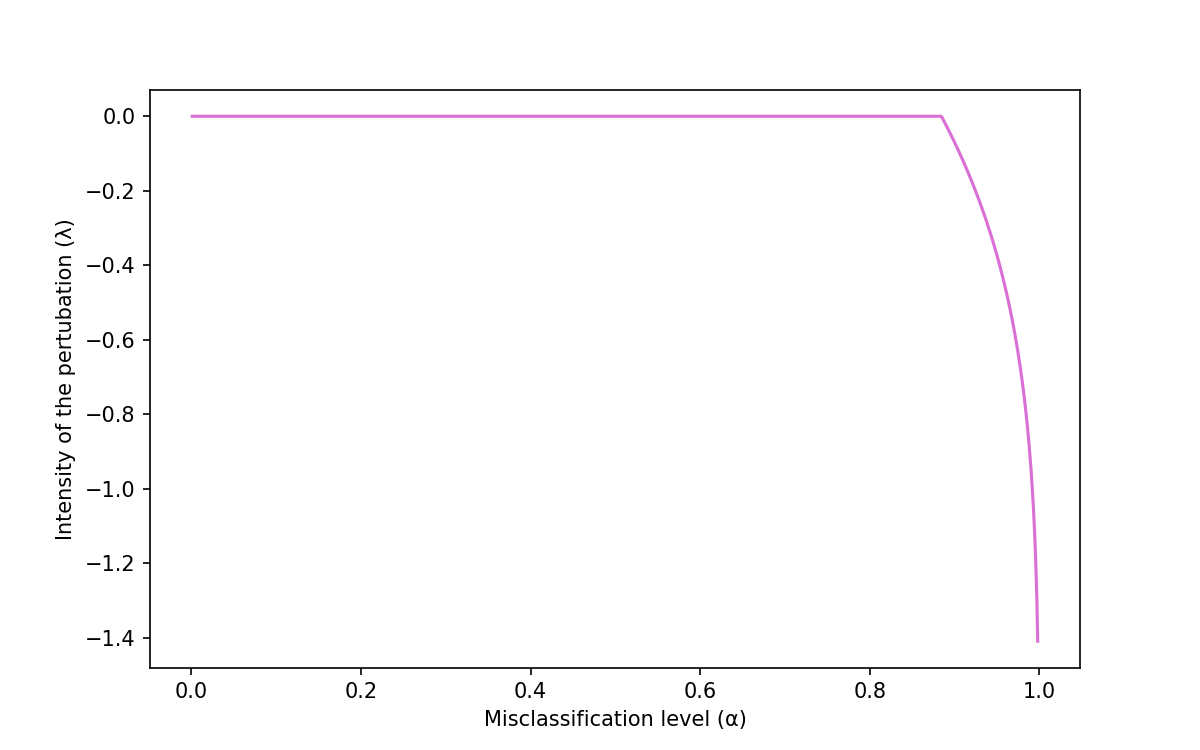}}
	\caption{Test Image 2}\label{subfig:cats_intensity_vs_alpha_test_b}
\end{subfigure}
\caption{Intensities of perturbations versus misclassification levels.} 
\label{fig:cats_intensity_vs_alpha_test}
\end{center}
\vskip -0.2in
\end{figure*}

We choose to perturb 2 squared images from the test set, which are represented in Figure \ref{fig:cats_adv_test}. Original and adversarial images cannot be distinguish by the human eye. 
Image 1 is correctly classified as a cat. The intensity of the perturbation of this image is a increasing function of the misclassification level (Figure~\ref{subfig:cats_intensity_vs_alpha_test_a}): as $\alpha$ increases, the associated adversarial examples is further away from the decision hyperplane. Image 2 is not correctly classified as a cat. Then, the perturbation intensity is negative and is a decreasing function of the misclassification level: a stronger misclassification implies to be further away from the decision boundary in the same half-space where is the original example. As seen in Figure \ref{subfig:cats_adv_test_b} and \ref{subfig:cats_intensity_vs_alpha_test_b}, the value of \(\lambda^*\) associated to Image 2 for \(\alpha=0.75\) is \(0\), because the probability that the original example is misclassified is \(0.88\).

\begin{figure}[tb]
%\vskip 0.2in %so everything fits in the page
\begin{center}
\centerline{\includegraphics[width=\columnwidth]{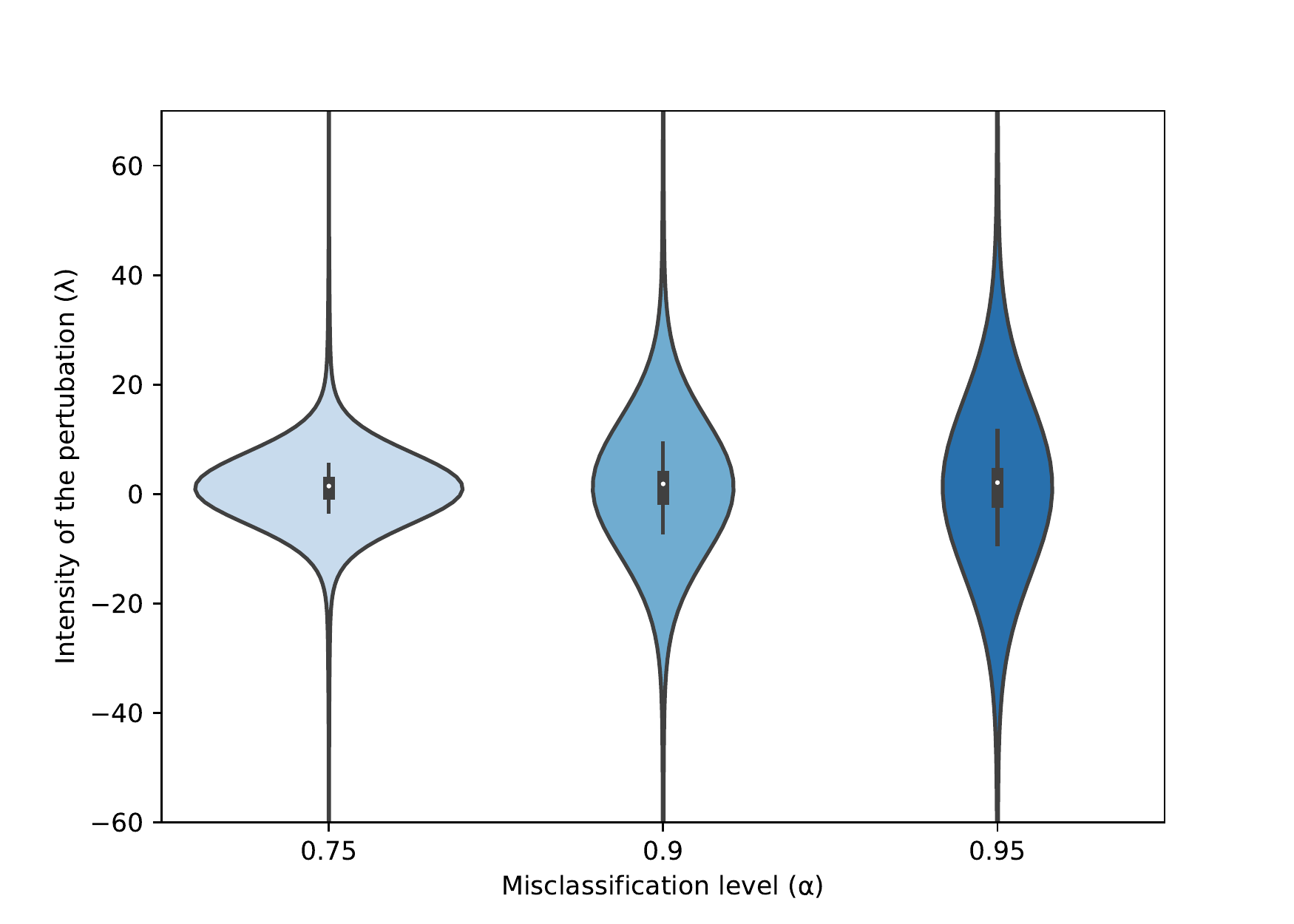}}
\caption{Violin plot of the intensities of perturbations in the test set for different levels of misclassification. The figure is zoomed in for better visualization.}
\label{fig:cats_violinplot}
\end{center}
\vskip -0.2in
\end{figure}

We computed the intensities associated to the misclassification levels of \(0.75\), \(0.9\) and \(0.95\), for each test example. The empirical distributions of \(\lambda^*\) grouped by \(\alpha\) are represented as violin plots in Figure \ref{fig:cats_violinplot}. The intensities are scattered, because of the differences in scales of the initial perturbations~\(\delta_0\), and the fact that the variance-covariance matrix of~$\beta$ is higher in some directions of \(\delta_0\) than others. Moreover, increasing the misclassification level seems to lead to higher empirical variance of the intensities.

\subsection{Conclusions, limitations and future
work}\label{conclusions-limitations-and-future-work}

In this paper, we show a simple way to craft an adversarial example that
achieves an expected misclassification rate in the case of limited
knowledge, in which the attacker knows that the defender uses a logistic
regression, but doesn't know the defender's training data. We defined an
adversarial example having an expected misclassification rate of
\(\alpha\) by the defender, as an \(\alpha\)-adversarial example. Using
2 real-world datasets, we show the importance to compute the intensity
of adversarial perturbations at the individual level: computing an
adversarial perturbation on the attacker surrogate model and applying
the same intensity across all perturbations is a suboptimal strategy to
achieve satisfactory intra-technique transferability.

Our method can be used on any adversarial perturbation technique that
only uses the surrogate attacker model without considering the
defender's model. But it is based on the assumptions that \emph{(i)} the
attacker has a very large number of training examples, \emph{(ii)} the
defender has a very large number of training data generated by the same
DGP than the defender's data, and \emph{(iii)} the specifications
(optimization method, regularization, hyperparameters, etc.) are known.
Moreover, to be computationally feasible, the number of features \(p\)
cannot be very large, because the computation of the variance-covariance
matrix of the parameters needs the inversion of a \(p \times p\) matrix.

Future research may be to:

\begin{itemize}
\item
  Extend our results to multinomial logistic regressions
\item
  Add other penalization methods 
\item
  Use finite sample distributions to have a better estimate of the
  variance-covariance matrix of the parameters when the number of
  training data is not large
\item
  Solve optimally the optimization problem \ref{eq:min_delta_star} to compute~\(\delta^*\)
  instead of the suboptimal solution \(\lambda^*\)
\item
  Handle the additional constraints listed in section \ref{introduction}, like
  \(x_0 + \delta \in \mathcal{X}\)
\item
  Extend our method to other models that have known asymptotic or
  finite-sample parameters distributions
\item
  Evaluate the cross-technique transferability of \(\alpha\)-adversarial
  examples
\item
  Extend the method of \(\alpha\)-adversarial examples to the case of
  unknown model, unknown model specification or unknown hyperparameters,
  but known distributions of these elements.
\end{itemize}

% do not put float after
\FloatBarrier

%\clearpage

%%%% Bibliography

\nocite{kaggle_dogs_nodate}
\nocite{lichman_uci_2013}
\nocite{seabold_statsmodels:_2010}
\nocite{pedregosa_scikit-learn:_2011}

\bibliography{adv_logistic}
\bibliographystyle{icml2018}

\end{document}